\documentclass{article}

\usepackage[final]{corl_2019} 
\usepackage[utf8]{inputenc} 
\usepackage[T1]{fontenc}    
\usepackage{url}            
\usepackage{booktabs}       
\usepackage{amsfonts}       
\usepackage{amsbsy}
\usepackage{nicefrac}       
\usepackage{microtype}      
\usepackage{subcaption}

\usepackage{natbib}
\usepackage{graphicx}
\usepackage{graphics}
\usepackage{multirow}
\usepackage{xcolor}
\usepackage{amsmath,commath}
\usepackage{float}
\usepackage[bottom]{footmisc}

\usepackage{algorithm}
\usepackage{algorithmic}

\DeclareMathOperator*{\atanh}{atanh}

\newcommand{\bE}{\mathbb{E}}

\usepackage{todonotes}

\newif\ifshowcomments

\showcommentstrue

\let\oldtodo\todo
\renewcommand{\todo}[1] {{\ifshowcomments{\oldtodo{#1}}\fi}}

\title{Continuous-Discrete Reinforcement Learning for Hybrid Control in Robotics}

\author{
  Michael Neunert*\\
  \texttt{neunertm@google.com}\\
  \And
  Abbas Abdolmaleki*\\
  \texttt{aabdolmaleki@google.com}\\
  \AND
  Markus Wulfmeier\\
  \And
  Thomas Lampe\\
  \And
  Jost Tobias Springenberg\\
  \And
  Roland Hafner\\
  \And
  Francesco Romano\\
  \And
  Jonas Buchli\\
  \And
  Nicolas Heess\\
  \And
  Martin Riedmiller\\
  \AND 
  ~\\
  \vspace{-2cm}
  DeepMind, United Kingdom
}

\begin{document}
\maketitle


\begin{abstract}
Many real-world control problems involve both discrete decision variables -- such as the choice of control modes, gear switching or digital outputs -- as well as continuous decision variables -- such as velocity setpoints, control gains or analogue outputs. However, when defining the corresponding optimal control or reinforcement learning problem, it is commonly approximated with fully continuous or fully discrete action spaces. 
These simplifications aim at tailoring the problem to a particular algorithm or solver which may only support one type of action space. Alternatively, expert heuristics are used to remove discrete actions from an otherwise continuous space. In contrast, we propose to treat hybrid problems in their `native' form by solving them with hybrid reinforcement learning, which optimizes for discrete and continuous actions simultaneously. In our experiments, we first demonstrate that the proposed approach efficiently solves such natively hybrid reinforcement learning problems. We then show, both in simulation and on robotic hardware, the benefits of removing possibly imperfect expert-designed heuristics. Lastly, hybrid reinforcement learning encourages us to rethink problem definitions. We propose reformulating control problems, e.g. by adding meta actions, to improve exploration or reduce mechanical wear and tear.
\end{abstract}



\section{Introduction}
	
Recent advances in numerical optimal control and Reinforcement Learning (RL) are enabling researchers to study increasingly complex control problems \citep{mnih2015human,openai2018dexterous,silver2017mastering,tassa2014control}. Many of these problems, both in simulation and the real world, have hybrid dynamics and action spaces, consisting of continuous and discrete decision variables. In robotics, common examples for continuous actions are analogue outputs, torques or velocities while discrete actions can be control modes, gear switching or discrete valves. Also, outside of robotics we find many hybrid control problems such as in computer games where mouse or joystick inputs are continuous but button presses or clicks are discrete. However, many state-of-the-art RL approaches have been optimized to work well with either discrete (e.g. \citep{mnih2015human}) or continuous (e.g. MPO \citep{abdolmaleki2018maximum, abdolmaleki2018relative}, SVG \citep{heess2015learning}, DDPG \citep{lillicrap2015continuous} or Soft Actor Critic \citep{haarnoja2018soft}) action spaces but can rarely handle both -- notable exceptions are policy gradient methods, e.g. \citep{schulman2017} -- or perform better in one parameterization than another \citep{openai2018dexterous}. 
This can make it convenient to transform all control variables so that they can be handled by the a single paradigm -- e.g by discretizing continuous variables, or by approximating discrete actions as continuous by thresholding them on the environment side instead of as part of the RL agent. Alternatively, control variables may be removed from the optimization problem e.g. by using expert-designed heuristics for discrete variables in continuous problems. Although either approach can work practice, in general, both strategies effectively reduce control authority or remove structure from the problem, which can affect performance or in the end make a problem harder to solve.

In this work, we propose to approach these problems in their native form, i.e. as hybrid control problems with partially discrete and partially continuous action spaces. We derive a data efficient model free RL algorithm that is able to solve control problems with both continuous and discrete action spaces as well as hybrid optimal control problems with controlled (and autonomous) switching. Being able to handle both discrete and continuous actions robustly with the same algorithm allows us to choose the most natural solution strategy for any given problem rather than letting algorithmic convenience dictate this choice. We demonstrate the effectiveness of our approach on a set of control problems with a focus on robotics tasks, both in simulation and on hardware. These examples contain native hybrid problems, but we also demonstrate how a hybrid approach allows for novel formulations of existing problems by augmenting the action space with `meta actions' or other quasi-hierarchical schemes. With minimal algorithmic changes this enables, for instance, variable rate control and thereby improving energy efficiency and reducing mechanical wear. More generally it allows to implement strategies that can address some of the challenges in RL such as exploration.

\section{Related Work}
\label{sec:related_work}


Control problems with continuous and discrete action spaces, in practice, each tend to have their own idiosyncrasies and challenges. In consequence, even though approaches exist that can, in principle, deal with either type of decision variable, different types of algorithms are favored for solving the different types of problems. To tackle hybrid reinforcement learning problems, recent work such as P-DQN \citep{xiong2018parametrized} and Q-PAMDP \citep{masson2016reinforcement} simply combine a discrete and a continuous RL algorithm and hence do not solve the problem in a unified fashion. A substantial part of hybrid RL literature focuses on a subcategory called Parameterized Action Space Markov Decision Processes (PAMDP)~\citep{masson2016reinforcement, hausknecht2015deep, bester2019multi, fan2019hybrid}, which is a hierarchical problem where the agent first selects a discrete action and subsequently a continuous set of parameters for that action. \citet{masson2016reinforcement} solve PAMDPs using a low dimensional policy parameterization. In contrast, \citep{hausknecht2015deep, baumann2018deep, fan2019hybrid} make use of continuous (deep) RL to output both continuous continuous actions and weights for the discrete choices. Subsequently, an \emph{argmax}-operation (or \emph{softmax}) is applied to select the discrete action to be applied to the system.  While our work can also be applied to PAMDPs, it is not limited to this special case but can solve general hybrid problems. Vice-versa, despite designed for solving PAMDPs, the `argmax-trick` used in \citep{hausknecht2015deep, fan2019hybrid} can be used to solve also non-hierarchical hybrid problems. Similarly, a range of work on hierarchical control in continuous action spaces effectively optimizes hybrid action spaces \cite{bacon2017option, wulfmeier2019regularized, zhang2019dac, harut2019termination}.

Alternatively to gradient-based optimization, evolutionary approaches~\citep{salimans2017evolution,back1996evolutionary,hansen2001completely} can be applied to address hybrid control problems. However, even recent approaches do not achieve a comparable data-efficiency \citep{salimans2017evolution}, which limits their use for low-data regimes such as robotics. In the optimal control literature, hybrid problems are frequently tackled using (discrete) Dynamic Programming, e.g. Value or Policy Iteration~\citep{branicky1998unified}. Other approaches include partitioning the system into continuous subsystems and then approaching the problem in a hierarchical fashion, i.e. solving each continuous control problem separately and adding a discrete controller on top (see surveys in \citep{bemporad1999control, branicky1998unified}) or tackling the overall problem using game theory~\citep{lygeros1995game}. Some approaches treat the entire system as continuous (see survey in \citep{branicky1998unified}). One can then try to solve the problem using state-of-the-art (continuous) numerical optimal control. However, most of these algorithms rely on differentiability of the system dynamics with respect to control actions, which is not the case for discrete actions. To overcome this issue, Mixed Integer Programming has been used~\citep{bemporad1999control} which solves an optimization problem involving both continuous and discrete optimization variables. With increase in computational power, Mixed Integer approaches allow for online or Model Predictive Control (MPC) implementations ~\citep{bemporad1999control}, often relying on locally linear-quadratic approximations. However, solving larger scale Mixed-Integer, especially in presence of constraints, non-differentiable costs or non-linearities, can still be very challenging due to the combinatorial complexity. 



\section{Preliminaries}
 
We consider reinforcement learning with an agent operating in a Markov Decision Process (MDP) consisting of the state space $\mathcal{S}$, a hybrid continuous and discrete action space $\mathcal{A}$, an initial state distribution $p(s_0)$, transition probabilities $p(s_{t+1} | s_t, a_t)$ which specify the probability of transitioning from state $s_t$ to $s_{t+1}$ under action $a_{t}$, a reward function $r(s, a) \in \mathbb{R}$ and the discount factor ${\gamma \in [0, 1)}$. The actions are drawn from a policy which is a state-conditioned probability distribution over actions $\pi(a | s)$. The objective is to optimize,
$\textstyle{J(\pi) = \mathbb{E}_{\pi,p(s_0),p(s_{t+1} | s_t, a_t)} \Big\lbrack \sum_{t=0}^\infty \gamma^t r(s_t, a_t) \Big\rbrack}$
where the expectation is taken with respect to the trajectory distribution induced by $\pi$.
We also define the action-value function for policy $\pi$ as the expected discounted return when choosing action $a$ in state $s$ and acting subsequently according to policy $\pi$ as $Q^\pi(s,a) = \mathbb{E}_\pi \lbrack \sum_{t=0}^\infty \gamma^t r(s_t, a_t) | s_0=s, a_0 = a]$. This function satisfies the recursive expression $Q(s_t, a_t) = \mathbb{E}_{p(s_{t+1} | s_t, a_t)} \big[ r(s_t, a_t) + \gamma V(s_{t+1}) \big]$ where $V^\pi(s) = \mathbb{E}_\pi[ Q^\pi(s,a) ]$ is the value function of $\pi$.

\section{Method}
In this section we introduce a class of policies for hybrid discrete-continuous decision making and required update rules for optimizing such policies. 
 
\subsection{Hybrid Policies}\label{sec:HybridPolicies}
We consider hybrid policies of a simple class in this paper which will allow us to represent action spaces with both continuous and discrete dimensions -- or purely continuous / discrete action spaces if desired. Formally, we define a hybrid policy $\pi(a | s)$ as a state dependent distribution that jointly models discrete and continuous random variables. We assume independence between action dimensions, denoted by $a^i$, for simplicity, i.e,
\begin{eqnarray}\label{eq:policy}
\pi_{\theta}(a | s) =  \pi^{c}_\theta(a^\mathcal{C} |s) \pi^{d}_\theta(a^\mathcal{D} |s) = \prod_{a^i \in a^\mathcal{C}} \pi^{c}_\theta(a^i|s) \prod_{a^i \in a^\mathcal{D}} \pi^{d}_\theta(a^i|s),
\end{eqnarray}
where $a^\mathcal{C}$ and $a^\mathcal{D}$ are the sub-sets of action dimensions with continuous values and discrete values respectively (with $\mathcal{C}$ and $\mathcal{D}$ representing continuous and discrete action spaces). 
We represent each component of the continuous policy $\pi^{c}_\theta$ as normal distribution 
$$
 \pi^{c}_\theta(a^i|s) = \mathcal{N}(\mu_{i,\theta}(s),\sigma^2_{i,\theta}(s)),
$$
and represent the discrete policy $\pi^{d}$ as a (per-dimension) categorical distribution over $K$ discrete choices parameterized by state-dependent probabilities $\alpha_\theta(s)$
$$
 \pi^{d}_\theta(a^i|s) = \textrm{Cat}^{i}(\alpha^{i}_{\theta}(s)) , \ \ \ \forall{i,s} \sum_{k=1}^K \alpha_{k,\theta}^i(s) = 1,
$$
where $\theta$ comprises the parameters of all policy components which we want to optimize. In particular, we will consider the case where $\alpha^{i}_{\theta}(s)$, $\mu_{i,\theta}(s),\sigma^2_{i,\theta}(s)$ are all represented as outputs of a neural network. We refer to the appendix for additional details on sampling and computing log probabilities of this policy parameterization.

\subsection{Hybrid Policy Optimization}\label{sec:method}
As pointed out in Section~\ref{sec:related_work}, 
a number of policy optimizers are in principle capable of optimizing hybrid policies.
In this work we build on top of the MPO algorithm \citep{abdolmaleki2018maximum,abdolmaleki2018relative}. MPO is a state-of-the-art policy optimization algorithm that allows us to use off-policy data and has been shown to be both data efficient and robust. 
To train a hybrid policy using MPO we rely on a learned approximation to the Q-function $\hat{Q}(s, a) \approx Q^\pi(s, a)$ (we refer to the appendix for an exposition of how this can be learned from a replay buffer $\mathcal{R}$). Using this Q-function, MPO updates the policy in two steps.

\textbf{Step 1:} first we construct a non-parametric improved policy $q$. This is done by maximizing $\bar{J}(s,q) = \mathbb{E}_{q(a|s)}[\hat{Q}(s,a)]$ for states $s$ drawn from a replay buffer $\mathcal{R}$ while ensuring that the solution stays close to the current policy $\pi_k$, i.e. $\mathbb{E}_{s \sim \mathcal{R}}[\text{KL}(q(a|s) \| \pi_{k}(a|s))] < \epsilon$, where $\mathrm{KL}$ denotes the Kullback-Leibler divergence. This optimization has a closed form solution given as $q(a\vert s)\propto \pi_k(a \vert s) \exp{\nicefrac{\hat{Q}(s,a)}{\eta}},$ where $\eta$ is a temperature parameter that can be computed by minimizing a convex dual function (\cite{abdolmaleki2018maximum}). 

\textbf{Step 2:} we use supervised learning to fit a new parametric policy to samples from $q(a|s)$. Concretely, to obtain the parametric policy, we solve the following weighted maximum likelihood problem with a constraint on the change of the policy from one iteration to the next 

\begin{equation}
\begin{aligned}
    \theta_{k+1} & = \arg \max_{\theta} \mathbb{E}_{s \sim \mathcal{R}}\Big[ \mathrm{KL}\big( q(a | s) \|  \pi^{c}_\theta(a^\mathcal{C} |s) \pi^{d}_\theta(a^\mathcal{D} |s)\big) \Big] \\
    \text{s.t. } & \mathbb{E}_{s \sim \mathcal{R}} \Bigg[ \mathrm{KL}(\pi^{c}_{\theta_k}(a^\mathcal{C} | s) \| \pi^{c}_\theta(a^\mathcal{C} | s))\Bigg] \!<\! \epsilon_c ,~~~ \mathbb{E}_{s \sim \mathcal{R}} \Bigg[\frac{1}{K}\sum_{i=1}^K  \mathrm{KL}(\pi^{d}_{\theta_k}(a^i|s)||\pi^{d}_\theta(a^i|s))  \Bigg]\! <\! \epsilon_d.
\end{aligned}
\label{eq:objective_pi}
\end{equation}
To control the change of the continuous and discrete policies independently, we take an approach similar to \cite{abdolmaleki2018relative} and decouple the optimization of discrete and continuous policies. This allows us to enforce separate constraints for each:
The first constraint bounds the $\mathrm{KL}$ divergence between the old and new continuous policy by $\epsilon_c$ and the second one bounds the average KL across categorical distributions by $\epsilon_d$. 
To solve Equation \eqref{eq:objective_pi}, we first employ Lagrangian relaxation to make it amenable to gradient based optimization and then perform a fixed number of gradient ascent steps (using Adam \citep{kingma2014adam}); details can be found in the Appendix.

\section{Verification and Application of Hybrid Reinforcement Learning}
\label{sec:result}
With a scalable hybrid RL approach we are no longer bound to formulating purely discrete or continuous problems. This enables us to solve native hybrid control problems in their natural form, but we can also reformulate control problems (or create entirely new ones) to address various issues in control and RL. These novel formulations can lead to better overall performance and learning speed but also reduce the amount of expert knowledge required. We group our experiments in three categories: The first set is used to verify the algorithm, the second set contains examples of `native` hybrid problems and finally we show novel hybrid formulations in continuous domains, adding `meta` control actions.

\subsection{Validating Hybrid MPO}
While there are established benchmarks and proven algorithms for continuous (and discrete) reinforcement learning, there are not many for hybrid RL. We run validation experiments on continuous domains where we partially or fully discretize the action space. This allows us to test our approach on well established benchmarks and compare it to state-of-the-art continuous control and hybrid RL algorithms. 
We consider the DeepMind Control Suite~\citep{tassa2018deepmind} which consists of a broad range of continuous control challenges, as well as a real world manipulation setup with a robotic arm.

\subsubsection{Comparison of Continuous, Discrete and Hybrid Control on Control Suite}
\label{sec:control_suite_compare}
For our comparison to a continuous baseline, we run Hybrid MPO in two settings, first fully discrete and second hybrid (partially discrete, partially hybrid). In the hybrid setting we discretize the last two action dimensions while in the fully discrete one we discretize all action dimensions. In both the hybrid and the discrete approach, we discretize the continuous actions coarsely, allowing only three distinct actions: $\{-1, 0, 1\}$. We use regular MPO as our continuous benchmark. Results show no noticeable performance difference between the fully continuous baseline and both the fully discrete as well as the hybrid approach. Surprisingly, despite the coarse control, both learning speed and final performance are practically identical (see Figure~\ref{fig:control_suite_comparison} in the Appendix for learning curves). Also, computational complexity is similar. Hence, we conclude that the approach is comparable in data efficiency to standard MPO and that our implementation is sound from an optimization perspective.

\subsubsection{Comparison to the `argmax-trick'}
In our second test, we compare Hybrid MPO against the `argmax-trick' proposed in the PAMDP literature~\citep{hausknecht2015deep, fan2019hybrid} representing the state-of-the-art in this field. The `argmax-trick' models discrete actions with continuous weights for each (action) option and subsequently applies the option with the highest weight (or samples from the softmax distribution). While developed for PAMDPs, it can be directly applied to non-hierarchical hybrid problems as well. Hence, we believe it is insightful to use a similar approach as a baseline to understand how well a purely continuous RL approach can scale to hybrid problems. For a fair comparison, and to eliminate any influence of the specific RL algorithm, we apply the `argmax-trick' structure to MPO (instead of using DDPG as in~\citep{hausknecht2015deep} or PPO as in~\citep{fan2019hybrid}). Effectively, we use fully continuous MPO with an \emph{argmax}-operation on continuous weights for each discrete action dimension. We refer to it as `argmax-MPO'.
In all experiments we are fully discretizing all action dimensions independently at regular intervals between the upper and lower action limit\footnote{As continuous dimensions are handled the same in all implementations, we discretize the full action space.}. We vary the resolution between 3 and 61 control values per dimension, exploring different trade offs between problem size and the granularity of control.
\begin{figure}[htbp]
    \centering
    \includegraphics[width=0.9\textwidth]{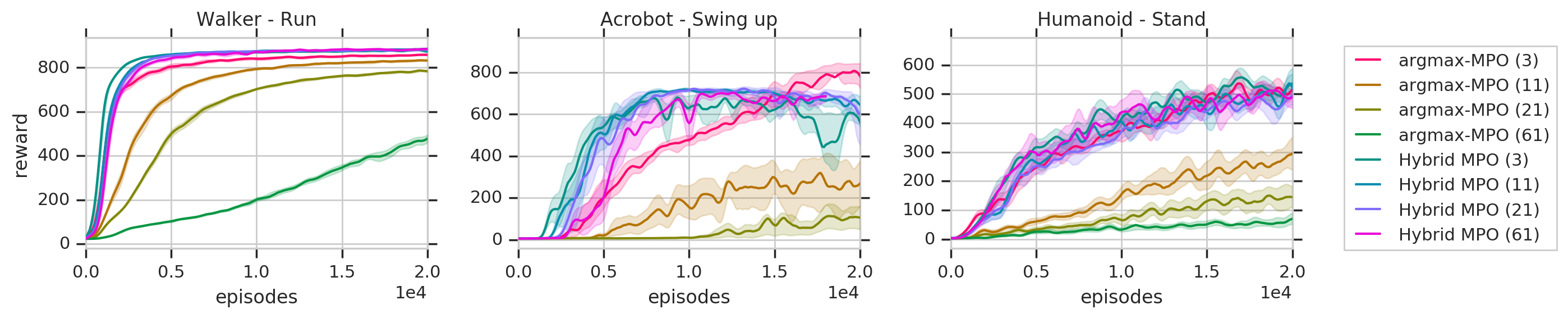}
    \caption{Hybrid MPO and argmax-MPO applied to a subset of the Control Suite. The number of discrete options for each action dimension is provided in braces.}
    \label{fig:control_suite_discrete}
\end{figure}
We observe that throughout all tasks, Hybrid MPO learns much faster. A subset of the results are provided in Figure~\ref{fig:control_suite_discrete}. Even when comparing fine resolution Hybrid MPO experiments with coarse argmax-MPO, it still trains faster despite the increase in parameters. There is little difference between both approaches in final performance, however it might take argmax-MPO significantly longer to reach it. argmax-MPO's learning speed scales quite poorly with task complexity and action resolution. These results are in line with the experiments in previous work~\citep{hausknecht2015deep, baumann2018deep} where the `argmax-trick' was applied to low dimensional problems with few discrete options. In comparison, Hybrid MPO is almost unaffected by the action resolution and no noticeable difference between 3 and 61 action discretization is visible. In tasks where finer control is required, such as the Humanoid task in Figure~\ref{fig:control_suite_discrete}, Hybrid MPO actually performs better with finer action resolution. Another big difference between Hybrid MPO and argmax-MPO is scaling of computational complexity. Hybrid MPO sees less than 15\% increase in computation time for the learning steps over the action discretizations, even for large action dimension tasks such as the humanoid. This also holds true for argmax-MPO in low dimensional tasks. However, in higher dimensional setups such as the humanoid, computation time (in our implementation) more than quadruples between 3 and 61 discretizations for the actions. On an absolute scale, Hybrid MPO outperforms argmax-MPO across all tasks and action resolutions, emphasizing the benefit of a natively hybrid implementation.

\subsubsection{Sawyer Reach-Grasp-Lift with Discrete Gripper Control}
\label{subsec:sawyer-lift}

\begin{figure}[htbp]
    \centering
    \includegraphics[height=0.23\textwidth]{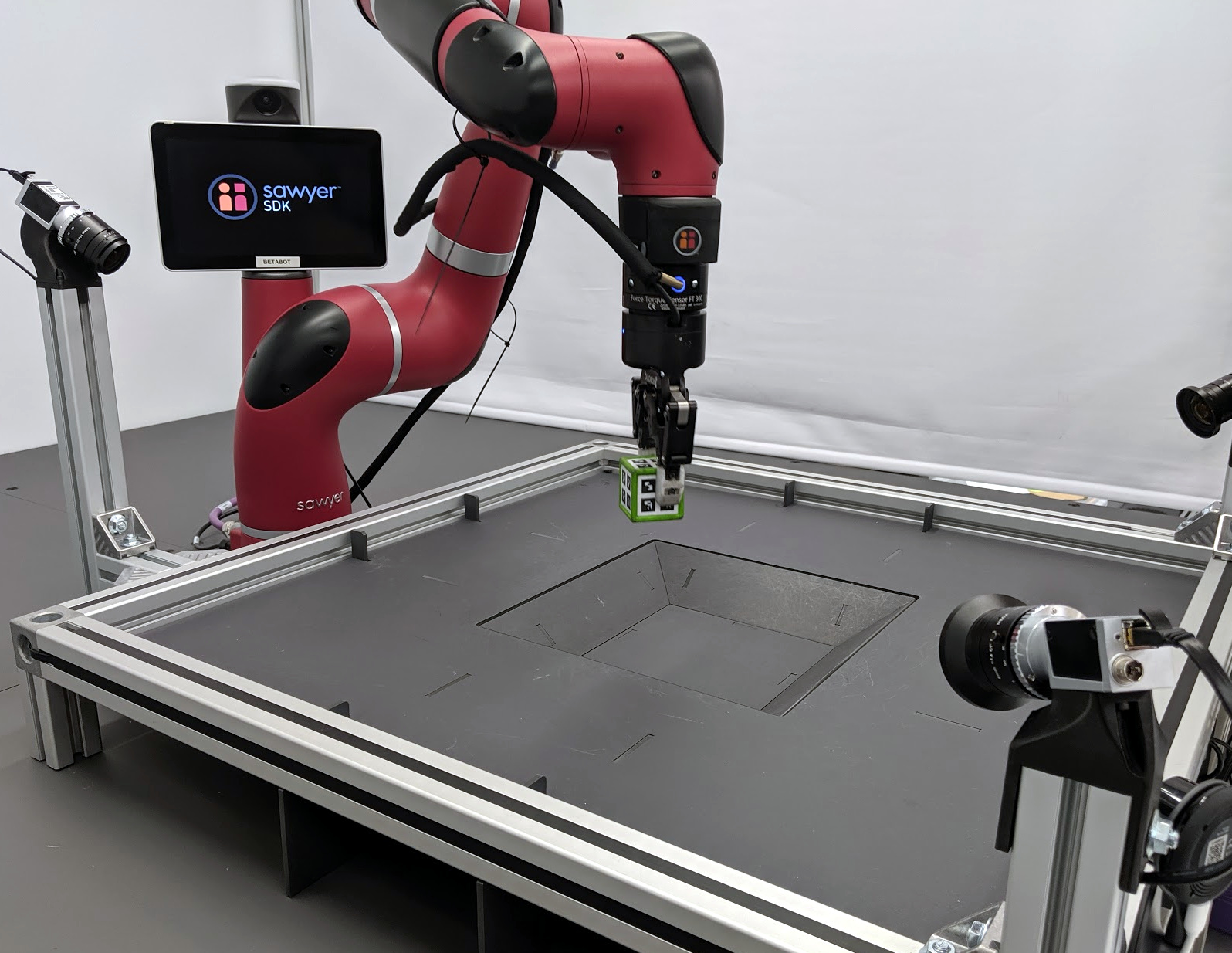}~~~
    \includegraphics[height=0.23\textwidth]{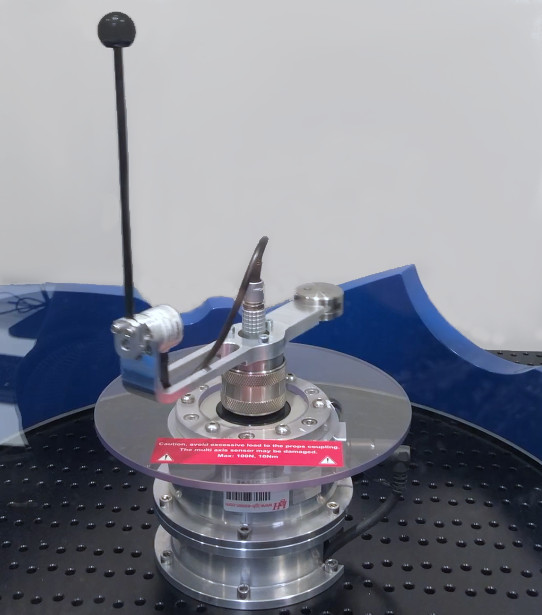}
    \caption{Sawyer cube manipulation setup (left) and Furuta Pendulum (right).}
    \label{fig:sawyer_furuta}
\end{figure}

In order to validate if the results above can be reproduced on hardware, we apply Hybrid MPO to a robotic manipulation task. We use Rethink Robotics Sawyer robot arm (for details on the setup, see Appendix~\ref{app:sawyer_setup}). The goal is to reach, grasp and lift a cube, as shown in Figure~\ref{fig:sawyer_furuta}. The reward is the sum of these three sub-tasks where the reach-reward is given for minimizing the distance between gripper and the cube, the grasp reward is a sparse, binary reward for triggering the built-in grasp sensor and lift is a shaped reward for increasing the height of the object above the table.
We run the baseline MPO algorithm with the arm and the gripper controlled in continuous velocity mode, while for Hybrid MPO we discretize the gripper velocity to $\{-1, 1\}$, i.e. open or close at full speed. The results in Figure~\ref{fig:sawyer_inverted_pendulum} left\footnote{We will revisit this setup in Section~\ref{sec:sawyer_action_repeat}. Hence the Figure contains learning curves not discussed yet.} show that the Hybrid MPO approach significantly outperforms the baseline, which is unable to solve the task. The reason lies in exploration: In order to reach the cube, the agent needs to open the gripper. However, to grasp the block the gripper needs to be closed again. Both tasks required opposite behaviour and, since the grasp reward is sparse and the gripper is slow, this poses a challenging exploration problem for learning to grasp. 
Initially, the Gaussian policy will have most of its probability mass concentrated on small action values and will thus struggle to move the gripper's fingers enough to see any grasp reward, explaining the plateau in the learning curve.
The Hybrid MPO approach on the other hand always operates the gripper at full velocity and hence exploration is improved -- allowing the robot to solve the task completely. While the improved exploration is a side effect of discretization, it underlines the shortcomings of Gaussian exploration. 

\subsection{Optimal Mode Selection for Peg-In-Hole}
\label{sec:peg_in_hole}
Many control problems are actually hybrid but are approximated as purely continuous or purely discrete. Examples are systems that combine continuous actions with a mode selection mechanism or discrete events. These choices are often excluded from the problem and the discrete choice/action is either fixed or selected based on heuristics. A common example in classical formulations is the choice of control mode or action space. Usually an expert chooses the mode that seems most suitable for the task but this choice is rarely verified. In the following example (which can also be interpreted as a PAMDP), we show that Hybrid MPO allows for exposing multiple ``modes'' to the agent, allowing it to select the mode and a continuous action for it based on the current observation. As a test setup, we create a peg-in-hole setup where the robot has to perform precision insertion. In our example, we provide the agent with two (emulated) control modes that could resemble a gearbox with two gears: A coarse Cartesian velocity controller with limits of 0.07 m/s and a fine one of 0.01 m/s. We further impose wrist force limits resulting in episode termination to protect the setup and encouraging gentle insertion. The reward is shaped and computed based on the known hole position and forward kinematics. We train a hybrid agent that can switch between both modes (``virtual gears'') as well as two continuous actions for either fixed mode.

\begin{figure}[htbp]
    \centering
    \includegraphics[width=0.65\textwidth]{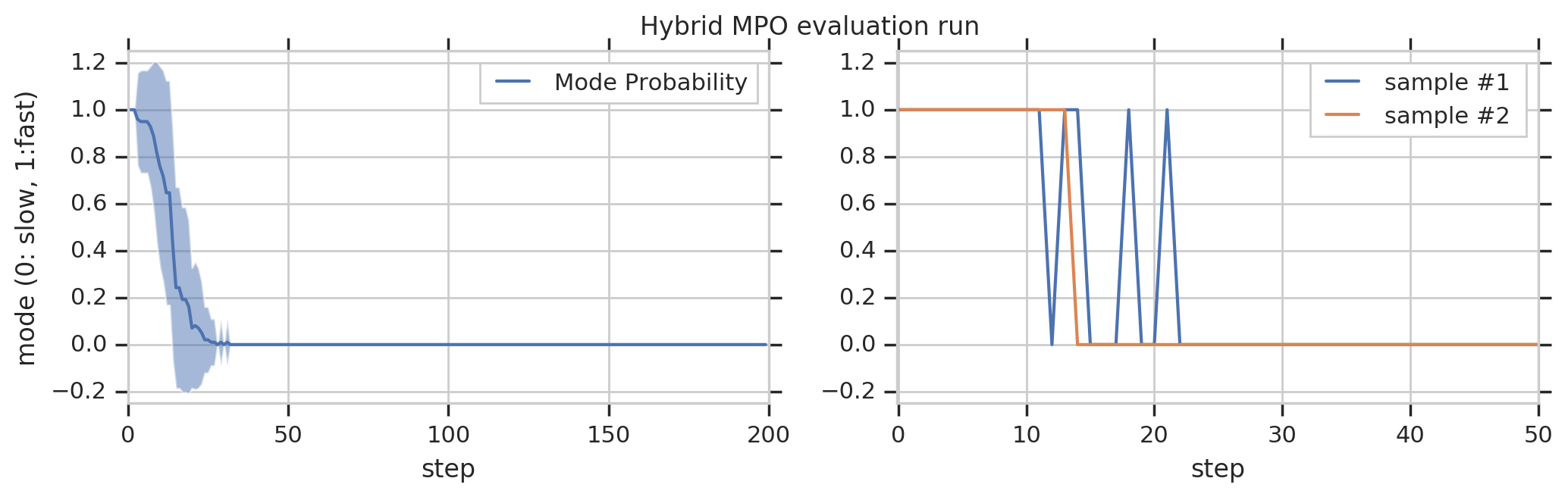}
    \caption{Peg-in-hole evaluation run for Hybrid MPO. The left plot shows the probability of the mode selection. The right hand plot shows the mode selection for two sample episodes (the mode remains zero, i.e. the slower control mode,for the steps 50-200 that are not shown).}
    \label{fig:evaluation}
\end{figure}

Our results show that the hybrid approach achieves an average final reward of approximately 2750 compared to 2500 for the fine control agent and around 1750 for the coarse mode agent. These results show that for an expert designer, selecting an appropriate mode beforehand can be difficult: The slow agent does not trigger the force limits but is much slower at inserting, leading to a lower average reward. The fast agent terminates frequently and while it might achieve good reward in some episodes, it cannot match the consistent performance of the hybrid agent, even after four times as much training time. While a switching heuristic could help alleviate the problem, designing an optimal heuristic is difficult and possibly a lengthy, iterative process.
In order to understand the mode switching behavior of the hybrid agent, we run 100 test evaluations. The left plot in Figure~\ref{fig:evaluation} shows the observed mean and standard deviation of the mode chosen by the Hybrid MPO approach counted over the 100 evaluation runs. The right plot in Figure~\ref{fig:evaluation} shows the agent's mode selection for two sample episodes, representative for all other episodes. The agent uses the coarse control mode to approach the hole quickly and consistently uses the fine action scale to perform the precise insertion. Some episodes show a ``wiggling'' behavior, where the Hybrid policy switches between modes near the peg while others show a single, discrete switch. While the experiment shows that an \emph{a priori} choice on the control mode can harm final performance and/or learning speed, one can even imagine tasks where a wrong mode selection would lead to complete failure of an experiment. There will also be tasks where the added complexity of choosing the mode negatively affects learning speed and using a single mode is optimal. But even in such cases, a hybrid approach is beneficial since it only requires a single experiment, whereas a fixed choice would still need to be verified by an ablation.

\subsection{Adding Meta Control to Continuous Domains}
The combination of discrete and continuous actions in Hybrid MPO opens up new ways of formulating control problems, even for fully continuous systems. Hybrid MPO can be used to add `meta actions', i.e. actions that modify the (continuous) actions or change the overall system behavior. In the following experiments, we demonstrate how such additional actions can e.g. improve exploration, solve event triggered control or reduce mechanical wear at training time.

\subsubsection{Furuta Pendulum}
\label{subsec:furuta}
The first set of experiments using `meta control actions' is conducted on a `Furuta Pendulum' (shown in Figure~\ref{fig:sawyer_furuta}) which is the rotational implementation of a cart pole. We define a sparse reward task that poses a hard exploration challenge: The reward is one when the pendulum is within a range of [-5, +5] degrees around the upright position and the main motor is in a range of [-15, 15] degrees around the backward pointing position. Otherwise the reward is zero. Before each episode, the motor is reset to the front facing position. Hence, in order to experience reward, first a large motion is required to move the main motor to the back and subsequent fast motions are required to swing the pendulum up. As a result, exploration without time correlation or with fixed time correlation will be quite poor. However, most RL approaches rely on such exploration. As a result, the baseline MPO agent struggles at solving the task. To improve exploration, we use Hybrid MPO and add a discrete `meta' action ``act-or-repeat'' to the problem, where the agent can choose to use the newly sampled continuous action or repeat the previous one (which is provided as an observation such that the problem remains an MDP)\footnote{This problem can be interpreted as a PAMDP where one discrete action has no continuous parameters.}. We bias the initial policy to choose ``repeat'' with a probability of 95\% to encourage exploration. Hence, there is stochastic action repeat leading to exploration at different frequencies. As results in Figure~\ref{fig:sawyer_inverted_pendulum} right show, the Hybrid MPO agent solves the task much quicker, despite having to learn to not repeat actions when balancing the pendulum. In the Appendix~\ref{app:furuta_all}, we also provide a comparisons on a sparse balancing task in front as well as a fully shaped task. These show that even in simple problems the additional ``act-or-repeat'' action does not harm (but also does not improve) learning speed or final performance.


\subsubsection{Sawyer Reach-Grasp-Lift with Agent-Controlled Action Repeat}
\label{sec:sawyer_action_repeat}
In this experiment, we test the improved exploration strategy on the robot arm block lifting task, described in Subsection~\ref{subsec:sawyer-lift}. While discretizing the gripper action helped with exploration, it is masking the underlying problem, which is a complex combination of small issues: in order to reach the cube without pushing it away, the agent learns to keep the gripper fingers open. However, the subsequent grasp is a sparse signal that requires closing the fingers by a fair amount. The gripper fingers are slow and thus they effectively act as a low pass filter that filters out most of the policy's (zero mean initialized) Gaussian exploration. Hence, with higher control frequency the issue becomes more severe, creating an unfavorable relation between control rate and exploration.
\begin{figure}[htbp]
    \centering
    \includegraphics[width=0.45\textwidth]{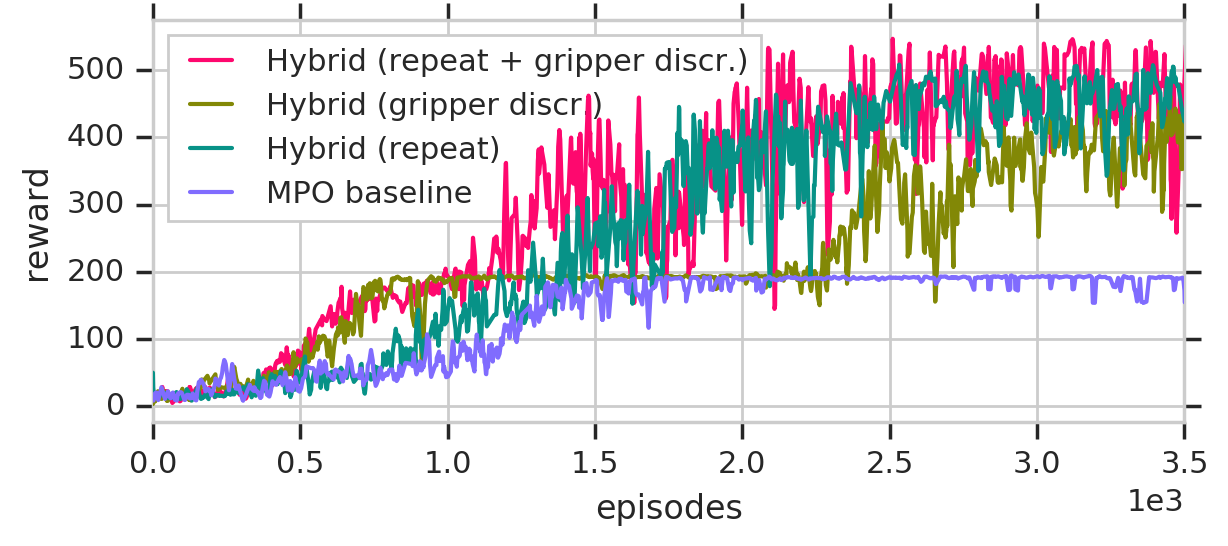}
    \includegraphics[width=0.45\textwidth]{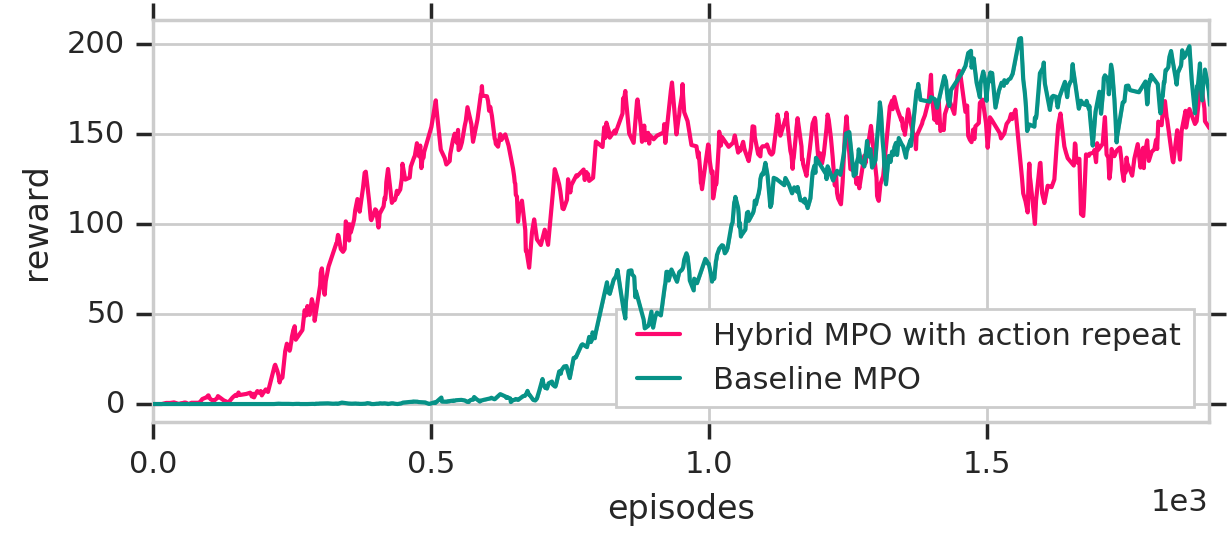}
    \caption{\textbf{Left}: Learning performance of Hybrid MPO with a discrete gripper action (ochre), an additional act-or-repeat action (green) and a combination of both (magenta) compared to the standard MPO implementation (blue) for an object lifting task. Standard MPO as well as the discrete gripper Hybrid MPO learn to reach quite quickly but then both suffers from poor exploration for learning to grasp. Hybrid MPO (with action repeats and/or discrete gripper) shows improved exploration behavior.
    \textbf{Right}: Learning curves for Furuta Pendulum experiments. The balancing in the back task with sparse reward underlines the exploration problem of the standard MPO algorithm.}
    \label{fig:sawyer_inverted_pendulum}
\end{figure}
By adding the same ``act-or-repeat'' action as in Subsection~\ref{subsec:furuta}, we can once again significantly improve exploration, as shown in Figure~\ref{fig:sawyer_inverted_pendulum} left. The Hybrid MPO agent with action repeat does not plateau after learning to reach since it experiences occasional reward from grasping and lifting early on in training. When comparing these results to the baseline, we can see that exploring at a single rate can lead to poor exploration, covering only a limited part of the state space, and effectively hinder learning.  When using an additional action to decide whether to repeat the previous (continuous) action or use the newly sampled action, we effectively explore at different rates. As in the Furuta Pendulum experiments, the additional action dimension does not seem to affect learning speed and final performance. We also run an experiment where we combine discrete gripper actions with our ``act-or-repeat'' action. Unsurprisingly, there is not much additional gain in learning speed or performance. However, there is a second side-effect of using action repeats: it leads to smoother exploration and fewer (mechanical) load direction changes. When dealing with hardware this significantly reduces wear and tear. One could also imagine encouraging action repeats as part of the reward function. This could be directly applied to Event Triggered Control~\citep{baumann2018deep}. The experiments also shows how exploration and control rate are often unnecessarily intertwined in RL. Larger step sizes lead to a different exploration behavior. However, exploration and control rate should not be coupled and Hybrid MPO (with action repeat) allows to disentangle both.

\subsubsection{Action Repeat on Control Suite Tasks}
We extend the experiments for agent controlled action repeats to the Control Suite domain. As the Furuta Pendulum results might already hint at, learning curves show no noticeable gain or performance drop from action repeats (see  Figure~\ref{fig:control_suite_comparison} in the Appendix). We assume that the improved exploration does not pay off in these tasks since rewards are strongly shaped, compared to the sparse rewards in the previous experiments. Given that action repeats do not seem to hurt performance, there is little reason not to make use of them.

\subsubsection{Action Attention on Control Suite Tasks}
\label{sec:action_attention_control_suite}
In the last application, we try to push the limits of Hybrid MPO. We test the agent in a setting that could be best described as ``action-attention'': From all available action dimensions, the agent can only affect one at a time while zero action is applied to all others. Hence, independent of the number of actuators ($n_{\text{act}}$) in the system, the agent has two actions: one discrete action, which can take any integer value in the range of $a_d = [0, n_{\text{act}}-1]$, and one continuous action between $a_c = [-1, +1]$, which is mapped to the action range of the particular actuator. Also this problem is consistent with the PAMDP formulation. We once again take a look at Control Suite tasks. Compared to the previous experiments where the agent could control all degrees of freedom simultaneously, the ``action-attention'' poses a severe limitation in control authority, especially with increasing degrees of freedom. Not unexpectedly, there is a loss in performance especially in high-dimensional domains. Yet, the agent copes with the control authority loss by learning alternative strategies which can be best seen in the supplemental video. One notable example is the ``swimmer'' task where the agent's action attention travels in wave form through the body, actuating one link after the other. A second example is the ``walker'' where the agent resorts to ``walking on its knees'' hence focusing its limited attention capacity on a small number of degrees of freedom rather than spreading it across all.

\section{Conclusion}
\label{sec:conclusion}
In this work, we have introduced a native hybrid reinforcement learning approach that can solve problems with mixed continuous and discrete action spaces. It can be applied both to hierarchical (PAMDPs) and non-hierarchical hybrid control problems. Our algorithm outperforms continuous-policy-based hybrid algorithms in general hybrid problems. Our experiments and application examples show the potential of Hybrid MPO to treat hybrid problems in their native form, improving performance, auxiliary criteria (such as controlling at variable rate) and removing expert priors. We further show that hybrid RL can be used to rethink the way we set up our control problems. By adding discrete `meta actions' to otherwise continuous problems, we can partially address common reinforcement learning pitfalls such as exploration or mechanical wear during training. While we provide a diverse set of examples, we believe there are many more applications of Hybrid RL that should be explored in future work. Finally, a detailed evaluation specifically for PAMDPs using benchmark problems and algorithms from the PAMDP literature could be insightful.



\acknowledgments{
The authors would like to thank the DeepMind robotics team for support for the robotics experiments.
}


\bibliography{example}  

\newpage
\appendix
\section*{Appendix}
\section{General Algorithm}
We use policy iteration procedure similar to MPO \cite{abdolmaleki2018maximum, abdolmaleki2018relative} for optimizing hybrid policies we are interested in. In order to update the policy, we first learn a Q-function. This step is also known as policy evaluation step. Subsequently, we use the Q-function to update the policy. This step is also known as policy improvement. In order to act in the environment we directly sample from the hybrid policy. We store the transitions and log probabilities of actions in the replay buffer. Next we explain how we sample from the hybrid policy and how we compute log probabilities.

\subsection{Sampling and Log probabilities}
As we showed in main paper we model the (behaviour) policy with a joint distribution of categorical and normal distributions (Eq. 1 main paper). And we use this policy to sample both continuous and discrete actions at the same time. This is done by sampling from normal and categoricals independently and concatenating them to obtain the desired action vector. In this case the log probability of the action vector is simply sum of the log probabilities under the normal and categorical distributions as we assumed independence in Eq. 1. We store the log probabilities of the actions that were taken in the replay buffer (i.e. the log probability of the behavior policy at that point in time) to learn the Q-function.

\subsection{Policy Evaluation}

As mentioned above, we need to have access to a Q-function to update the policy. While any method for policy evaluation can be used, we rely on the Retrace algorithm \citep{munos2016safe}. 
Concretely, we fit the Q-function $Q(s, a, \phi)$. Note that in this paper $a$ is sampled from the hybrid policy and it is a vector with both discrete and continuous dimensions. $Q(s, a, \phi)$ is represented by a neural network, with parameters $\phi$ which is obtained by minimising the squared loss:

\begin{equation}
\begin{aligned}
&\min_\phi L(\phi) = \min_\phi \bE_{\mu_b(s), b(a|s)} \Big[ \big( Q(s_t, a_t, \phi) - Q^{\text{ret}}_t \big)^2 \Big], \text{with } \\ 
  &Q^{\text{ret}}_t = Q_{\phi'}(s_t, a_t)+ \sum_{j=t}^N \gamma^{j-t} \Big( \prod_{k=t+1}^j c_k \Big) \Big[ r(s_j, a_j) + \bE_{\pi(a | s_{j+1})} [ Q_{\phi'}(s_{j+1}, a) ] - Q_{\phi'}(s_j, a_j) \Big], \\
  &c_k = \min\Big(1, \frac{\pi(a_k | s_k)}{b(a_k | s_k)}\Big),
\end{aligned}
\end{equation}

where $Q_{\phi'}(s, a)$ denotes the target network for Q-function, with parameters $\phi'$, that we copy from the current parameters $\phi$ after a fixed number of steps. $N$ denotes number of steps for reward accumulation before we bootstrap with $Q_{\phi'}$. Additionally, $b(a|s)$ denotes the probabilities of an arbitrary behaviour policy. In our case we use an experience replay buffer and hence $b$ is given by the action probabilities stored in the buffer; which correspond to the action probabilities at the time of action selection.

\subsection{Maximum a Posteriori Policy Optimization for Hybrid Policies}
Given the Q-function, and current policy $\pi(a|s,\theta_k)$, in each policy improvement step MPO performs an EM-style procedure for policy optimization. In the E-step a sample based optimal policy $q(a|s)$ is found by maximizing: 

\begin{equation}
  \begin{aligned}
  & \max_q \bE_{\mu(s)} \Big[ \bE_{q(a|s)} \Big[ Q(s,a) \Big] \Big] \\
  & s.t. \bE_{\mu(s)} \Big[ \textrm{KL}(q(a|s) , \pi(a|s,\theta_k) ) \Big] <
  \epsilon.
  \end{aligned}
  \label{eq:e_step_hard}
\end{equation}
 
where $\mu(s)$ is the state distribution. In our case state distribution is represented by state samples in the experience replay buffer. We can solve the above equation and obtain the sample-based $q$ distribution in closed form,

\begin{equation}
q(a|s) \propto \pi(a|s,\theta_k)\exp\Big( \frac{Q(s,a)}{\eta^{*}}\Big),
\end{equation}
where we can obtain $\eta^{*}$ by minimising the following convex dual function,
\begin{equation}
  g(\eta) = \eta \epsilon + \eta \int \mu(s) \log \int \pi(a|s, \theta_k)\exp\Big( \frac{Q(s,a)}{\eta}\Big) \dif a \dif s,
  \label{eq:dual}
\end{equation}

see MPO paper \cite{abdolmaleki2018maximum} for more details on this optimization procedure. Note that this step only needs samples from the current policy and is not dependent on parametric form of the policy. Afterwards the parametric policy is fitted via weighted maximum likelihood learning (subject to staying close to the old policy) given via the objective:

\begin{equation}
\begin{aligned}
\theta_{k+1} & = \arg \max_{\theta}  \bE_{\mu(s)} \Big[ \bE_{q(a|s)} \Big[ \log \pi(a|s) \Big] \Big] \\
\text{s.t.} \ & \bE_{\mu(s)} \Big[ \textrm{KL}(\pi_{k}(a|s) , \pi(a|s) ) \Big] < \epsilon.
\end{aligned}
\label{eq:m_step_constraint}
\end{equation}

As in this paper we assume a hybrid continuous-discrete policy $\pi^{c}_\theta(a^\mathcal{C} |s) \pi^{d}_\theta(a^\mathcal{D} |s)$  , 
this objective can further be decoupled into continuous and discrete parts for the constraint which allows for  independent control over the change of continuous and discrete policies:

\begin{equation}
\begin{aligned}
    \theta_{k+1} & = \arg \max_{\theta} \mathbb{E}_{\mu(s)}\Big[\mathbb{E}_{q(a|s)}\Big[ \log\big( \pi^{c}_\theta(a^\mathcal{C} |s) \pi^{d}_\theta(a^\mathcal{D} |s)\big) \Big]\Big] \\
    \text{s.t. } & \mathcal{T}_c < \epsilon_c , \ \ \ \mathcal{T}_d < \epsilon_d,
\end{aligned}
\label{eq:objective_pi2}
\end{equation}

where $$\mathcal{T}_c = \mathbb{E}_{s \sim \mu(s)} \Bigg[ \mathrm{KL}(\pi^{c}_{\theta_k}(a^\mathcal{C} | s) \| \pi^{c}_\theta(a^\mathcal{C} | s))\Bigg]$$ and $$\mathcal{T}_d = \mathbb{E}_{s \sim \mu(s)} \Bigg[\frac{1}{K}\sum_{i=1}^K  \mathrm{KL}(\pi^{d}_{\theta_k}(a^i|s)||\pi^{d}_\theta(a^i|s))  \Bigg]. $$

The first constraint bounds the KL divergence of the continuous policy, which is a Gaussian distribution (as in this paper), over continuous dimensions and second constraint bounds the average KL divergence across all $K$ discrete dimensions where each dimension is represented by a categorical distribution. To solve this constrained optimisation we first write the generalised Lagrangian equation, i.e,

$$L(\theta,\eta_{c},\eta_{d}) = \mathbb{E}_{\mu(s)}\Big[\mathbb{E}_{q(a|s)}\Big[ \log\big( \pi^{c}_\theta(a^\mathcal{C} |s) \pi^{d}_\theta(a^\mathcal{D} |s)\big) \Big]\Big] + \eta_{c}(\epsilon_{c} - \mathcal{T}_{c}) + \eta_{d}(\epsilon_{d} - \mathcal{T}_{d})$$

Where $\eta_{c}$ and $\eta_{d}$ are Lagrangian multipliers.
And we solve the following primal problem,

$$\max_{\theta}\min_{\eta_{c}>0,\eta_{d}>0} L(\theta,\eta_{c},\eta_{d}).$$

In order to solve for $\theta$ we iteratively solve the inner and outer optimisation programs independently: We fix the Lagrangian multipliers to their current value and optimise for $\theta$ (outer maximisation) and then fix the parameters $\theta$ to their current value and optimise for the Lagrangian multipliers (inner minimisation). We continue this procedure until policy parameters $\theta$ and Lagrangian multipliers converge.

\section{Hyperparameters and Reward Functions}
\subsection{Hyperparameters}
All experiments are run in a single actor setup and we use the same hyperparameters for both MPO and Hybrid MPO (where they exist for both).
Tables \ref{t:MPO} show the hyperparameters we used for the experiments.

\begin{table}[t]
\begin{center}
 \begin{tabular}{c||c} 
 Hyperparameters &  \\
 \hline
 Policy net & 200-200-200\\ 
 Number of actions sampled per state& 20\\
 Q function net & 500-500-500\\
 $\epsilon$ & 0.1 \\
 $\epsilon_{d}$ & 0.0001 (0.01 on hardware) \\
 $\epsilon_{c}$ & 0.001\\
 Discount factor ($\gamma$) & 0.99 \\
 Adam learning rate & 0.0003 \\
 Replay buffer size & 2000000 \\
 Target network update period & 250\\
 Batch size & 3072\\
 Activation function & elu\\
 Layer norm on first layer & Yes\\
 Tanh on output of layer norm & Yes\\
 Tanh on Gaussian mean & No \\
 Min variance & Zero\\
 Max variance & unbounded \\
 Max transition use & 500
\end{tabular}
\end{center}
\caption{Hyperparameters for the proposed algorithm}
\label{t:MPO}
\end{table}

\subsection{Reward Functions}
\subsubsection{Control Suite}
All Control Suite tasks are run with their default reward functions.

\subsubsection{Sawyer Cube Manipulation}
The reward function for cube manipulation is defined as
\begin{equation}
    r(s_t,a_t) = \frac{1}{3}\text{reach} + \frac{1}{3}\text{grasp} + \frac{1}{3}\text{lift}
\end{equation}
where the individual components are defined as
\begin{align}
    \text{reach} &= 1 - \tanh^2( \frac{\atanh(\sqrt{0.95})}{0.1} |d_r|)\\
    \text{grasp} &=
    \begin{cases}
       \text{1,} &\quad\text{if grasp}\\
       \text{0}  &\quad\text{otherwise.} \\ 
     \end{cases} \\
    \text{lift}  &= 1 - \tanh^2( \frac{\atanh(\sqrt{0.95})}{h_{\text{max}}} |h_c-h_{\text{max}}|)
\end{align}
where $d_r$ is the Euclidean distance between the gripper's pinch position and the object position and $h_c$ is the height of the cube above the table and $h_\text{max}$ is the maximum height of the cube at 0.15 m.

\subsubsection{Furuta Pendulum}
The sparse reward function is defined as 
\begin{equation}
    r(s_t, a_t) = (|\alpha - 180^\circ| < 15^\circ) (|\beta|<5^\circ)
\end{equation}
where $\alpha$ is the main motor angle (measured from the front position) and $\beta$ is the pendulum angle (measured from the upright position).

The shaped reward is defined as

\begin{equation}
    r(s_t,a_t) = r_\text{pos} \cdot r_\text{vel} \cdot r_\text{action}
\end{equation}
where the individual terms are defined as
\begin{align}
     r_\text{pos} &= 1 - \tanh^2( \frac{\atanh(\sqrt{0.95})}{0.1} |e_\text{pos}|) \\
     r_\text{vel} &= \exp(-0.05 \norm{v} ) \\
     r_\text{action} &= \exp(-0.1|a_t|)
\end{align}
where $e_\text{pos} = 1 - \cos(\alpha)$ is the pendulum angle error and $v = [\dot{\alpha}, \dot{\beta} ]^T$ are motor and pendulum velocities.

\section{Hardware Setup}
\subsection{Sawyer Manipulation Setup}
\label{app:sawyer_setup}
The Sawyer cube manipulation setup consists of a Sawyer robot arm developed by Rethink Robotics. It is equipped with a Robotiq 2F85 gripper as well as a Robotiq FTS300 force torque sensor at the wrist. In front of the robot, there is a basket with a base size of 20x20 cm and inclined walls. The robot is controlled in Cartesian velocity space with a maximum speed of 0.07 m/s. The control mode has four control inputs: three Cartesian linear velocities as well as a rotational velocity of the gripper around the vertical axis. Together with the gripper finger angles, the total size of the action space is five.

Inside of the workspace of the robot is a 5x5x5 cm sized cube that is equipped with Augmented Reality markers on its surface. The markers are tracked by three cameras (two in front, one in the back). Using an extrinsic calibration of the cameras, the individual measurements are then fused into a single estimate which is provided as input to the agent. For the peg-in-hole experiments, the insertion is measured based on the known fixed position of the hole. Hence, no additional observation is provided to the agent.

The agent is run at 20 Hz and receives proprioception observations from the arm (joint and endeffector positions and velocities), the gripper (joint positions, velocities and grasp signal) and force-torque sensor (force and torques). In the case of cube experiments, the agent also receives the cube pose as an input. 

During all experiments, episodes are 600 steps long but are terminated early when the force-torque measurements of the wrist exceed 15 N on any of the principal axes.

\subsection{Furuta Pendulum}
The Furuta pendulum is driven by a single central brushless motor while the pendulum joint is passive and its position is measured with an encoder. The main motor is controlled in velocity mode and the agent controls the speed setpoint for the velocity controller.

The agent receives both the main motor and the pendulum's positions and velocities as observations. All position measurements are mapped to sine/cosine space to ensure any wrap around is handled gracefully and input to the agent is bounded.

All trained agents run at 100 Hz and episodes are 1000 steps long, unless the pendulum velocity is too high and the episode is terminated to protect the pendulum joint.

\section{Additional Experimental Results}
In this section, we provide a few more details on the experiments conducted as well as the learning curves. The main results are discussed in the paper but we decided to provide more details for the interested reader and for the sake of completeness and reproducability.

\subsection{Control suite comparison}
\label{app:control_suite}
\begin{figure}[tbp]
    \centering
    \includegraphics[width=\textwidth]{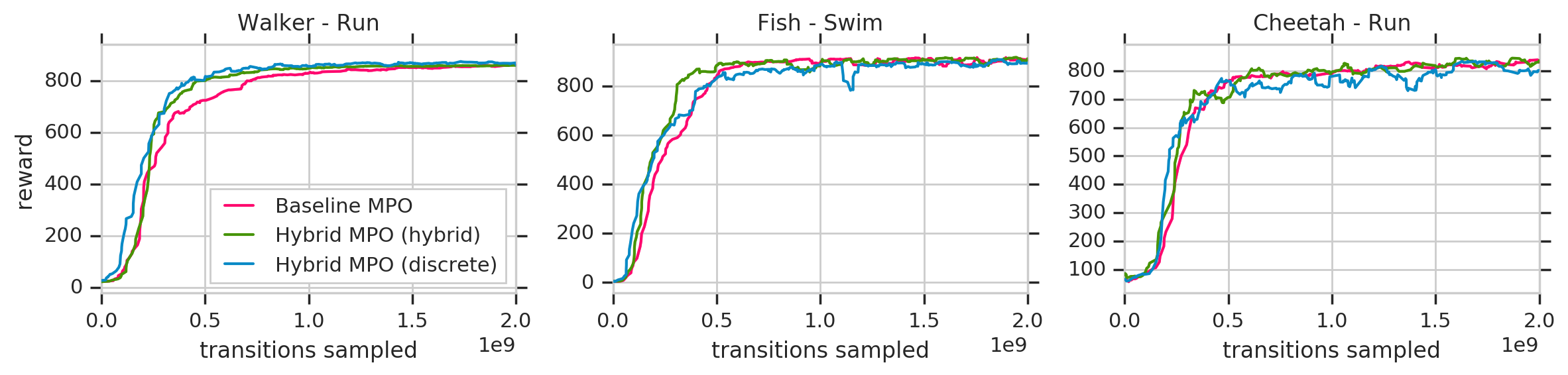}
    \caption{Comparison of learning curves for partially discrete and fully discrete Hybrid MPO to the baseline MPO implementation. The results show that Hybrid MPO performs comparable to the baseline in both settings. Hence, we assume that discrete actions do not significantly effect data efficiency or final performance.}
    \label{fig:control_suite_comparison}
\end{figure}

As described in Section~\ref{sec:control_suite_compare} we compare baseline MPO and hybrid MPO on partially and fully discretized versions of the Control Suite tasks. Figure~\ref{fig:control_suite_comparison} shows learning curves for a (representative) subset of the tested tasks. We did not see any significant changes in learning speed or final performance between the approaches. One reason that the same performance can be reached in a fully discrete setting is that the Control Suite tasks do not require extreme control precision and mechanical (rigid body dynamics) systems serve as some form of low pass filter, smoothing out discrete actions.

\subsection{Action attention}
\begin{figure}[tbp]
    \centering
    \includegraphics[width=\textwidth]{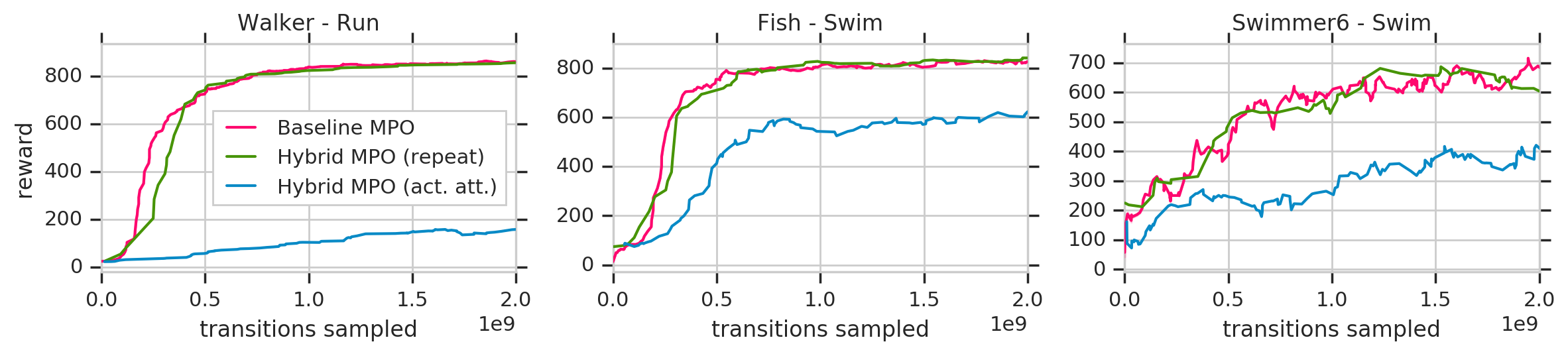}
    \caption{Comparison of learning curves for different variants of Hybrid MPO to the baseline MPO implementation. The results show that Hybrid MPO with action repeats performs comparable to the baseline. This result is not unexpected, given the strong shaping of the tasks. One difference though is the action attention setting, where the agent can only control one degree of freedom at a time. Given the loss in control authority, Hybrid MPO suffers in final performance but still comes up with interesting solutions to the problem (see video attachment).}
    \label{fig:control_suite_attention}
\end{figure}

In Section~\ref{sec:action_attention_control_suite} we describe the mechanism of `action attention' that we tested on the DeepMind Control Suite. Figure~\ref{fig:control_suite_attention} shows the learning curves for three tasks. As previously mentioned, the action attention agent does not (and possibly cannot, given the loss of control authority) reach the same performance as the baseline agent. However, it still learns and finds a solution that generates reward, occasionally coming up with novel strategies such as reducing the number of actuators the agent actually makes use of or coordinating them in sequence.

\subsection{Furuta Pendulum}
\label{app:furuta_all}
In Section~\ref{subsec:furuta}, we described the use of agent controlled action repeats and how they can be used for challenging exploration tasks such as balancing the Furuta Pendulum in the back using sparse rewards. Apart from this challenging task, we also tested two simpler tasks: Balancing the pendulum in front using sparse rewards as well as balancing the pendulum anywhere with shaped reward. The goal of these experiments were to verify that there is no performance loss in simple tasks by using (and having to `unlearn') action repeats.

\begin{figure}[htbp]
    \centering
    \includegraphics[width=0.8\textwidth]{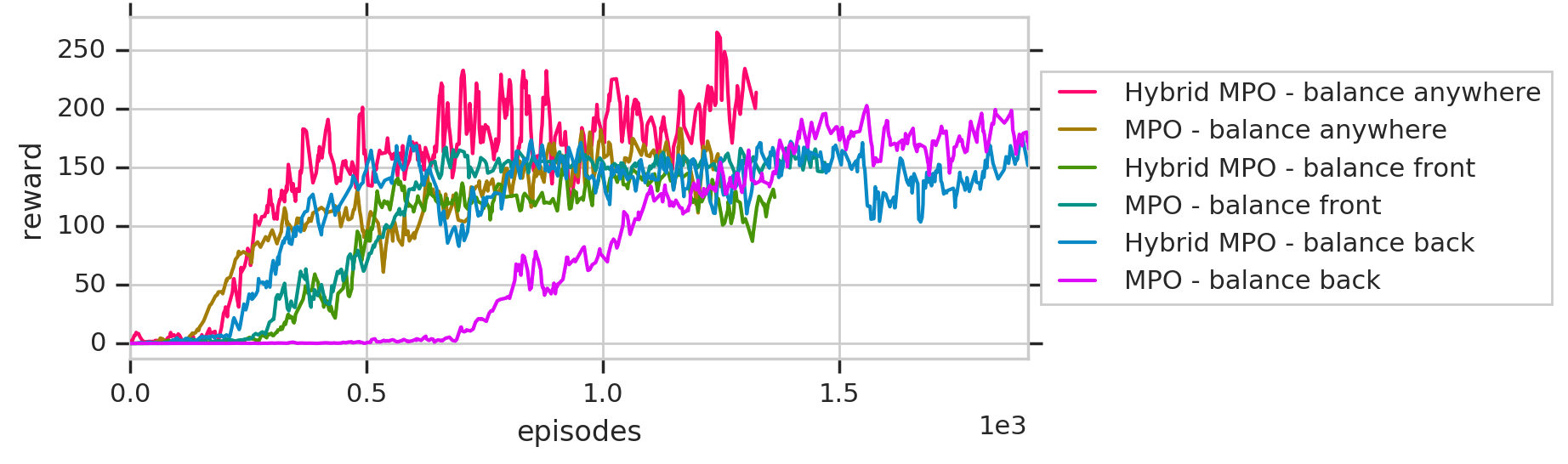}
    \caption{Learning curves for Furuta Pendulum experiments. There is little difference between the different tasks and approaches, demonstrating that action repeats in Hybrid MPO do not hurt final performance and the agent learns to control (`act') at a high rate. The balancing in the back task with sparse rewards underlines the exploration problem of the standard MPO algorithm while the action repeat in the Hybrid MPO approach does not result in any performance loss, also not in the easier tasks.}
    \label{fig:inverted_pendulum_all}
\end{figure}

As Figure~\ref{fig:inverted_pendulum_all} shows, performance and learning speed are not affected by the action repeat and both the hybrid as well as the baseline agent learn equally fast and reach the same performance. Hence, it seems as if there is no drawback in using action repeats by default, at least not in lower dimensional systems.

\subsection{Peg-In-Hole}
\begin{figure}[htbp]
    \centering
    \includegraphics[width=0.6\textwidth]{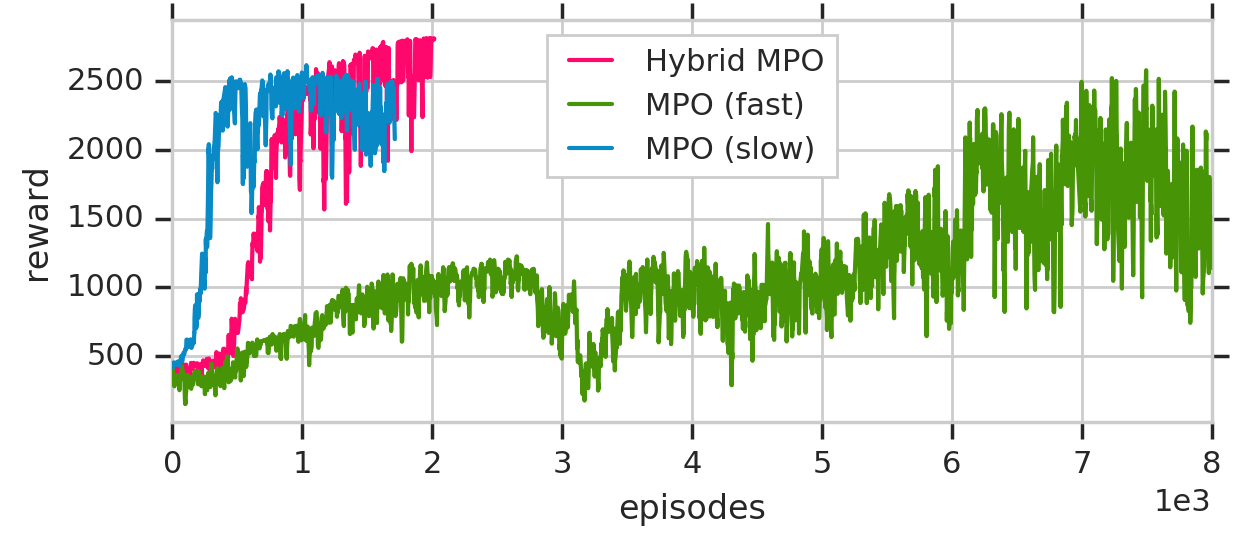}
    \caption{Learning curve for the peg-in-hole experiment. The red curve shows the Hybrid MPO agent which can switch between a fast and a slow mode and thus is able to successfully insert the peg. The green baseline is the standard MPO algorithm with only the fast control mode which suffers from many early episode termination due to collisions and hence is not able to insert the peg in early training. Blue is the MPO baseline with only the slow control mode. While it learns fast due to no early episode terminations, the final performance is lower due to the limited velocity lengthening the insertion process.}
    \label{fig:peg_in_hole_learning}
\end{figure}

Section~\ref{sec:peg_in_hole} discussed the `peg-in-hole' insertion tasks and provided basic performance figures. Figure~\ref{fig:peg_in_hole_learning} shows the learning curves for the same experiment. As one can observe, the coarse agent trains very slowly since it terminates episodes early due to exceeding the force/torque limits. The fine control agents trains fast, even faster than the hybrid one. However, it never reaches the same performance as the hybrid one. One might argue that an expert could come up with a mode switching strategy. However, it is unclear when the mode switch should happen, e.g. is it distance based and if so, at what distance shall it switch? So even this simple example shows that expert heuristics can be tricky to derive.

To put the learning curves into perspective, we evaluate the final policy for 100 episodes of 200 steps and compare it to the fast MPO baseline. During evaluation, we only apply the mean of the policy. We define success as having achieved full reward during at least one timestep in the episode. Table~\ref{tab:peg-in-hole-eval} summarizes the results. Hybrid MPO triggers the force-torque limits once and thus is unable to complete the task. Despite four times more training time, the fast baseline triggers the limits 8 times and additionally does not succeed in 3 episodes where the limits are not triggered.

\begin{table}[htbp]
\centering
\begin{tabular}{c|r|r}
\multicolumn{1}{l|}{}            & \multicolumn{1}{c|}{\textbf{Hybrid MPO}} & \multicolumn{1}{c}{\textbf{MPO (coarse)}} \\ \hline
early episode terminations       & 1/100                                    & 8/100                                       \\
successful episodes (max reward) & 99/100                                   & 89/100                                     
\end{tabular}
\vspace{0.3cm}
\caption{Peg-in-hole evaluation run statistics}
\label{tab:peg-in-hole-eval}
\end{table}

\end{document}